# 一种改进的 FastEuler-DLKF 小型无人机航姿算法研究

**摘　要**：精确的航姿参考系统(Attitude Heading Reference System, AHRS)是无人机可靠飞行系统的关键环节，本文针对小型无人机近地导航的应用场景，建立陀螺仪/加速度计/磁力计松耦合误差模型，提出一种改进的快速欧拉角双层卡尔曼滤波算法(FastEuler, Double-Layer Kalman Filter)。利用 MEMS 惯性测量单元(IMU)和磁力计等低成本器件，构建小型无人机航姿参考硬件和软件系统，搭建算法的离线和实时验证平台，分别在仿真和飞行测试环境下分析无人机姿态角变化。与此同时，对于加速度计中线性加速度对滚转角和俯仰角造成的有害影响，采用自适应因子对测量噪声协方差进行调节处理。通过与互补滤波算法进行实验对比表明，该算法可以在无人机飞行时提供准确的姿态角信息，提高了姿态解算的精确性和可靠性，消除了陀螺仪偏差对姿态估计的影响。

# An improved FastEuler-DLKF small-UAV AHRS algorithm

**Abstract**：The accurate Attitude Heading Reference System(AHRS) is an important apart of the UAV reliable flight system. Aiming at the application scenarios of near ground navigation of small-UAV, this paper establishes a loose couple error model of the gyroscope/accelerometer/magnetometer, and presents an improved FastEuler Double-Layer Kalman Filter algorithm. Using low-cost devices which include MEMS Inertial Measurement Units(IMU) and magnetometers, this paper constructs the AHRS hardware and software systems of UAV, and designs the offline and real-time verification platforms. Moreover, the attitude changes of UAV is analyzed by the simulation and flight test, respectively. In addition, an adaptive factor is used to adjust the measurement noise covariance in order to eliminate the harmful effects of linear acceleration in the accelerometer, which is solved the roll and ptich angle. The experimental comparison with the Complementary Filter shows that the proposed algorithm can provide accurate attitude information when UAV is flying, which improves the accuracy and reliability of attitude solution, and removes the influence the gyro bias for the attitude estimation.

## 0 引　言

通常情况下，航姿参考系统[1]是由陀螺仪、加速度计、磁力计和微处理器组成，并使用信息融合算法进行姿态角解算的航姿组合系统，而且在无其他传感器数据进行辅助的情况下依旧能够输出稳定的载体姿态信息。

目前对于 AHRS 算法的研究，主要包括互补滤波、梯度下降滤波以及卡尔曼滤波等。互补滤波是利用多个传感器的信号频率相互补偿特性，从频域上来处理噪声，融合不同的传感器数值来进行姿态解算。Mahony[4]提出一种互补滤波航姿算法，通过一阶动力学模型，融合低成本 IMU 和磁力计，为无人机提供良好的姿态估计参数。Jin Wu[5] 提出一种基于四元数航姿估计的快速互补滤波融合算法，并设计逐步滤波来消除磁力计异常值影响，在解算精度和实时性方面具有高效性。除了互补滤波，梯度下降滤波也常常用于航姿计算。梯度下降法是一种迭代求极值的算法，具体实现过程是按目标函数斜率的负方向来搜寻最优。Madgwick[6]将梯度下降法用于航姿解算，旨在用于人体运动追踪系统，使用四元数进行姿态更新，并进行磁失真补偿，使用加速度计和磁力计数据进行最优梯度下降分析，以计算陀螺仪测量误差的方向，并作为四元数导数。吕印新[7]针对低成本 MEMS 惯性器件，研究了一种基于梯度下降算法的四元数互补滤波算法，通过梯度下降法构造互补滤波四元数形式的反馈修正量，实现对四元数误差的校正和陀螺漂移的估计。

卡尔曼滤波[8]是一种从与目标信号相关的测量中，通过相关算法估计出目标信号的方法。在具有高斯分布噪声的系统中，常常作为一种最佳估计器，

通讯作者. E-mail: □□□□.

被用于目标追踪、航姿解算以及导航应用中。针对无人机航姿测量系统，宋宇[11]提出一种四元数卡尔曼滤波算法，设计自适应滤波修正测量噪声方差矩阵，不但解决了 MEMS 器件误差较大的问题，而且减少了陀螺仪随机误差对姿态估计的影响。Simone Sabatelli[12]使用一种九轴的惯性测量单元，提出一种四元数两级卡尔曼航姿滤波，第一级使用加速度数据，第二级使用磁力计数据，减少了传感器测量数据之间的耦合性，使得滤波解算更加灵活和安全。

本文在分析和总结常用航姿融合滤波算法的基础上，结合层级滤波和自适应因子调节的思想，提出一种 FastEuler-DLKF 无人机航姿滤波算法，使用姿态角误差和陀螺仪偏差作为状态量，用四元数进行姿态更新，提高了算法的实时运算效率和解算精度。

整体结构如下：首先建立了 IMU/磁力计航姿组合模型，根据无人机在近地导航的应用场景，对误差模型进行了合理简化；其次将加速度计和磁力计解算的欧拉姿态角作为测量值，并对异常姿态角进行剔除，对航向角进行修正。设计了 FastEuler-DLKF 航姿滤波融合框架，使用自适应因子对加速度计测量噪声方差进行调节，减少线性加速度对滚转角和俯仰角解算的影响；接着设计了无人机硬件和软件系统，利用搭建的实验平台采集飞行传感器数据，进行离线仿真实验和实时飞行测试实验；最后对算法进行分析讨论。

## 1 IMU/磁力计航姿组合建模

### 1.1 陀螺仪误差模型

当无人机在静止状态时，零偏误差是 MEMS 陀螺仪输出精度的主要影响因素，消除零偏误差可以提高陀螺仪三轴角速度测量精度。但是在无人机飞行时，陀螺仪误差中[13]的随机漂移误差会随着时间变化，因此需要对其进行建模估计。

$$\boldsymbol{\varepsilon}_g = \boldsymbol{\varepsilon}_0 + \boldsymbol{\varepsilon}_r + \boldsymbol{w}_\varepsilon \tag{1}$$

式(1)中，$\boldsymbol{\varepsilon}_0$ 为常值误差部分；$\boldsymbol{\varepsilon}_r$ 为随机漂移误差部分，$\boldsymbol{w}_\varepsilon$ 为陀螺仪误差中的白噪声，随机漂移误差部分可以表示为：

$$\boldsymbol{\varepsilon}_r = -\frac{1}{\tau_g}\boldsymbol{\varepsilon}_r + \boldsymbol{w}_{\varepsilon r} \tag{2}$$

式(2)中，$\tau_g$ 为一阶马尔科夫相关时间常数，$\boldsymbol{w}_{\varepsilon r}$ 为白噪声。由于 $\boldsymbol{\varepsilon}_0$ 在陀螺仪启动之后为常值，可以推导出：

$$\boldsymbol{\varepsilon}_0 = 0 \tag{3}$$

$$\boldsymbol{\varepsilon}_0 + \boldsymbol{\varepsilon}_r = \boldsymbol{\varepsilon}_r \tag{4}$$

可以将常值误差考虑进随机漂移误差内，因此陀螺仪误差模型如下：

$$\boldsymbol{\varepsilon}_g = \boldsymbol{\varepsilon}_r + \boldsymbol{w}_\varepsilon \tag{5}$$

### 1.2 姿态误差模型

捷联惯性姿态误差矢量方程[14]如下：

$$\delta\boldsymbol{\varphi} = \delta\boldsymbol{\varphi} \times \boldsymbol{\omega}_{in}^n + \delta\boldsymbol{\omega}_{in}^n - \boldsymbol{\varepsilon}_g^n \tag{6}$$

其中，$\delta\boldsymbol{\varphi}$ 为无人机姿态角误差，$\boldsymbol{\omega}_{in}^n$ 为导航系相对惯性系的角速度，$\boldsymbol{\varepsilon}_g^n$ 为导航坐标系中的陀螺仪误差。由于本文选用的 IMU 是 MEMS 低成本器件，相比高精度惯性器件而言，测量值噪声较大，一些小状态变化会淹没在测量噪声中；同时姿态误差方程的状态估计数值都是小量。式(6)中的误差模型对于研究嵌入式传感器航姿组合算法有些复杂，为了降低计算复杂度，提高算法更新频率，可对误差模型做一些合理简化。

由于航姿融合算法更新频率比较高，无人机位置变化引起的导航坐标系旋转可以忽略；地球旋转角速率相对于陀螺仪的测量精度也比较小，在姿态误差简化模型中可以对地球旋转角速率近似为 0。另外小型无人机的飞行范围在几公里到几十公里之间，可以忽略地球半径和曲率对航姿估计影响。因此由式(2)、式(5)以及式(6)可以得出简化姿态误差模型：

$$\delta\boldsymbol{\varphi} = -\boldsymbol{\varepsilon}_g^n = -C_b^n\boldsymbol{\varepsilon}_r^b - C_b^n w_\varepsilon \tag{7}$$

### 1.3 航姿组合模型

通过对陀螺仪误差模型和姿态误差模型的分析和简化，建立航姿误差线性高斯状态空间模型。

$$\begin{cases} x(t) = F(\mathrm{t})x(\mathrm{t}) + w(t) \\ \mathrm{z}(t) = H(\mathrm{t})x(t) + v(t) \end{cases} \tag{8}$$

式(8)中，$x(t)$ 为状态量，$F(\mathrm{t})$ 为状态转移矩阵，$w(t)$ 为过程噪声，$\mathrm{z}(t)$ 为量测量，$H(\mathrm{t})$ 为量测矩阵，$v(t)$ 为量测噪声。其中，假设 $w(t)$ 和 $v(t)$ 均是零均值高斯白噪声且互不相关。根据 1.1 和 1.2 小节误差模型可以得到航姿组合状态方程。

$$\begin{cases} \delta\boldsymbol{\varphi} = -C_b^n\boldsymbol{\varepsilon}_r^b - C_b^n\boldsymbol{w}_\varepsilon \\ \boldsymbol{\varepsilon}_r^b = -\dfrac{1}{\tau_g}\boldsymbol{\varepsilon}_r^b + \boldsymbol{w}_{\varepsilon r} \end{cases} \tag{9}$$

航姿组合量测方程用式(10)表示，其中 $\phi_a$ 和 $\theta_a$ 为加速计得到的滚转角和俯仰角，$\psi_m$ 为磁力计得到的航向角，$\hat{\phi}$、$\hat{\theta}$ 和 $\hat{\psi}$ 为陀螺仪四元数姿态更新估计值，$v_1(t)$ 为加速度计测量噪声，$v_2(t)$ 为磁力计测量噪声。

$$z(t) = \begin{cases} \begin{bmatrix} \phi_a - \hat{\phi} \\ \theta_a - \hat{\theta} \end{bmatrix} + v_1(t) \\ (\psi_m - \hat{\psi}) + v_2(\mathrm{t}) \end{cases} \tag{10}$$

## 2 FastEuler-DLKF 航姿滤波算法

### 2.1 陀螺仪四元数姿态更新

无人机陀螺仪旋转角速率动态响应比较迅速，在采样周期内，旋转角速率为可以看作为常值。也就是在 $\Delta t = t_k - t_{k-1}$ 时间内，$\boldsymbol{\omega}^b$ 是不变数值。由于陀螺仪采样频率比较高，更新时间间隔 $\Delta t$ 比较小，等效旋转矢量 $\boldsymbol{\Phi}$ 可以认为是比较小的量。因此，可以将等效旋转矢量使用 $\Delta t$ 时间内的角增量表示：

$$\begin{aligned}\boldsymbol{\Phi} = \boldsymbol{\omega}^b \Delta t &= \left[\omega_x^b, \omega_y^b, \omega_z^b\right]^T \Delta t \\ &= \left[\Delta\theta_x, \Delta\theta_y, \Delta\theta_z\right]^T\end{aligned} \tag{11}$$

$$\Phi = |\boldsymbol{\Phi}| = \Delta\theta = \sqrt{\Delta\theta_x^2 + \Delta\theta_y^2 + \Delta\theta_z^2} \tag{12}$$

将旋转矢量用单位四元数表示如下：

$$q(\Delta t) = \begin{bmatrix} \cos\dfrac{\Delta\theta}{2} \\ \dfrac{\Delta\theta_x}{\Delta\theta}\sin\dfrac{\Delta\theta}{2} \\ \dfrac{\Delta\theta_y}{\Delta\theta}\sin\dfrac{\Delta\theta}{2} \\ \dfrac{\Delta\theta_z}{\Delta\theta}\sin\dfrac{\Delta\theta}{2} \end{bmatrix} \tag{13}$$

根据四元数乘法规则，姿态更新可以用式(14)表示，其中 $\otimes$ 为四元数乘法符号。

$$q(t_k) = q(t_{k-1}) \otimes q(\Delta t) \tag{14}$$

### 2.2 FastEuler 算法

在航姿解算过程中，加速度计和磁力计测量值修正姿态角，不过直接用三轴加速度值和三轴磁强测量数值进行修正会增加计算量，而且当出现异常测量值不容易进行检测。由于本文航姿解算的应用场景是小型无人机近地导航，飞行速度相对较慢，提出快速欧拉角算法，将三轴加速度计和三轴磁力计数值解算成姿态角，并且对异常的姿态角进行剔除。

**算法 1：FastEuler**

**输入：三轴加速度计和三轴磁力计数值**

$\text{accel} = [ax \;\; ay \;\; az] \qquad \text{mag} = [mx \;\; my \;\; mz]$

**输出：** $\phi_a \;\; \theta_a \;\; \psi_m$

**Step 1：计算滚转角和俯仰角测量值**

**If** $\text{accel} \neq 0$ **and** $\left|\sqrt{ax^2 + ay^2 + az^2} - g\right| \leq \partial$ **then**

$$\begin{cases} \phi_a = \text{atan2}(-ay, -az) \\ \theta_a = \text{atan2}(ax, -az) \end{cases}$$

**End if**

**Step 2：计算偏航角测量值**

**If** $\text{mag} \neq 0$ **then**

$$\begin{cases} hx = mx\cos\theta_a + my\sin\theta_a\sin\phi_a + mz\sin\theta_a\cos\phi_a \\ hy = my\cos\phi_a - mz\sin\phi_a \end{cases}$$

$\psi_m = \text{atan2}(-hy, hx)$

**Step 3：判断偏航角航向**

**If** $\psi_m < 0$ **then**

$\psi_m = \psi_m + 2\pi$

**End if**

**End if**

### 2.3 DLKF 滤波算法

通过 1.3 小节航姿组合模型的建立，使用卡尔曼滤波[15]对传感器进行融合计算，由于 IMU 和磁力计等传感器测量数据采集频率不一致，因此本文提出一种双层卡尔曼滤波的思想，可以使得不同更新速率的传感器数值被触发时，进入滤波器进行姿态估计处理。DLKF 滤波算法如下：

$$\begin{cases} \hat{X}_{k/k-1} = \boldsymbol{\Phi}_{k/k-1}\hat{X}_{k-1} \\ P_{k/k-1} = \boldsymbol{\Phi}_{k/k-1}P_{k-1}\boldsymbol{\Phi}_{k/k-1}^T + Q_{k-1} \end{cases} \tag{15}$$

$$\begin{cases} K_k^1 = P_{k/k-1}H_k^{1,T}(H_k^1 P_{k/k-1} H_k^{1,T} + R_k^1)^{-1} \\ \hat{X}_k^1 = \hat{X}_{k/k-1} + K_k^1\left[Z_k^1 - H_k^1\hat{X}_{k/k-1}\right] \\ P_k^1 = \left(I - K_k^1 H_k^1\right)P_{k/k-1} \end{cases} \tag{16}$$

$$\begin{cases} K_k^2 = P_{k/k-1}H_k^{2,T}(H_k^2 P_{k/k-1} H_k^{2,T} + R_k^2)^{-1} \\ \hat{X}_k^2 = \hat{X}_{k/k-1} + K_k^2\left[Z_k^2 - H_k^2\hat{X}_{k/k-1}\right] \\ P_k^2 = \left(I - K_k^2 H_k^2\right)P_{k/k-1} \end{cases} \tag{17}$$

式(15)为 DLKF 滤波时间更新方程，式(16)为 DLKF 滤波第一层量测更新方程，式(17)为 DLKF 滤波第二层量测更新方程。

状态量 $X = [\delta\phi \;\delta\theta \;\delta\psi \;\boldsymbol{\varepsilon}_{rx}^b \;\boldsymbol{\varepsilon}_{ry}^b \;\boldsymbol{\varepsilon}_{rz}^b]^T$ 分别为姿态角误差以及陀螺仪偏差。对 1.3 小节状态方程进行离散化可得。

$$\boldsymbol{\Phi}_{k/k-1} = I + dt \cdot F(\text{t}) = \begin{bmatrix} I & -C_b^n dt \\ I & -\dfrac{1}{\tau_g}dt \end{bmatrix} \tag{18}$$

式(18)中，$C_b^n$ 为机体系到导航系的方向旋转矩

阵，$dt$ 为 IMU 更新速率。量测量 $Z_1=[\delta\phi_z \ \ \delta\theta_z]^T$ 和 $Z_2=\delta\psi_z$ 分别为加速度计和磁力计测量姿态角误差。对 1.3 小节的量测方程进行离散化可得。

$$\begin{cases} H_1=[I_{2\times2} \quad 0_{2\times2}] \\ H_2=[I_{1\times1} \quad 0_{1\times1}] \end{cases} \tag{19}$$

式(18)中 $H_1$ 和 $H_2$ 分别是加速度计和磁力计量测矩阵。

### 2.4 滤波初值选取与修正

对于航姿组合滤波初值选取，式(20)分别是滤波状态、协方差初值、加速度计量测噪声和磁力计量测噪声，式(21)是滤波过程噪声，其中噪声数值大小可以参考传感器数据手册参数表[13]。

$$\begin{cases} X_0=0 \\ P_0=I \end{cases} \quad R_a=\begin{bmatrix} 0.5 & 0 \\ 0 & 5 \end{bmatrix} \quad R_m=5 \tag{20}$$

$$Q=\begin{bmatrix} 0.1 & 0 & 0 & 0 & 0 & 0 \\ 0 & 0.1 & 0 & 0 & 0 & 0 \\ 0 & 0 & 0.1 & 0 & 0 & 0 \\ 0 & 0 & 0 & 0.01 & 0 & 0 \\ 0 & 0 & 0 & 0 & 0.01 & 0 \\ 0 & 0 & 0 & 0 & 0 & 0.01 \end{bmatrix} * 10^{-4} \tag{21}$$

虽然小型无人机飞行速度相对较慢，但是由本文采用加速度计解算滚转角和俯仰角，当无人机的线运动加速度不为 0 时，将会存在误差，且误差随着无人机线运动加速度增大而增大。因此对加速度计量测噪声协方差采用自适应因子进行调节。

$$R_a(t_k)=\gamma_a^2\begin{bmatrix} 0.5 & 0 \\ 0 & 5 \end{bmatrix} \tag{22}$$

$$\gamma_a^2=\lambda_a\left(\left|\ \|a(k)\|-\|g\|\ \right|\right) \tag{23}$$

式(23)中，$\lambda_a$ 为设定的权重因子，$a(k)$ 为加速度数值，$g$ 为当地重力加速度。当无人机飞行速度不断增大时，载体线运动加速度越大，加速度计对应的量测噪声协方差也越大，相应的第一层滤波增益矩阵将减小，提高卡尔曼滤波的估计精度。

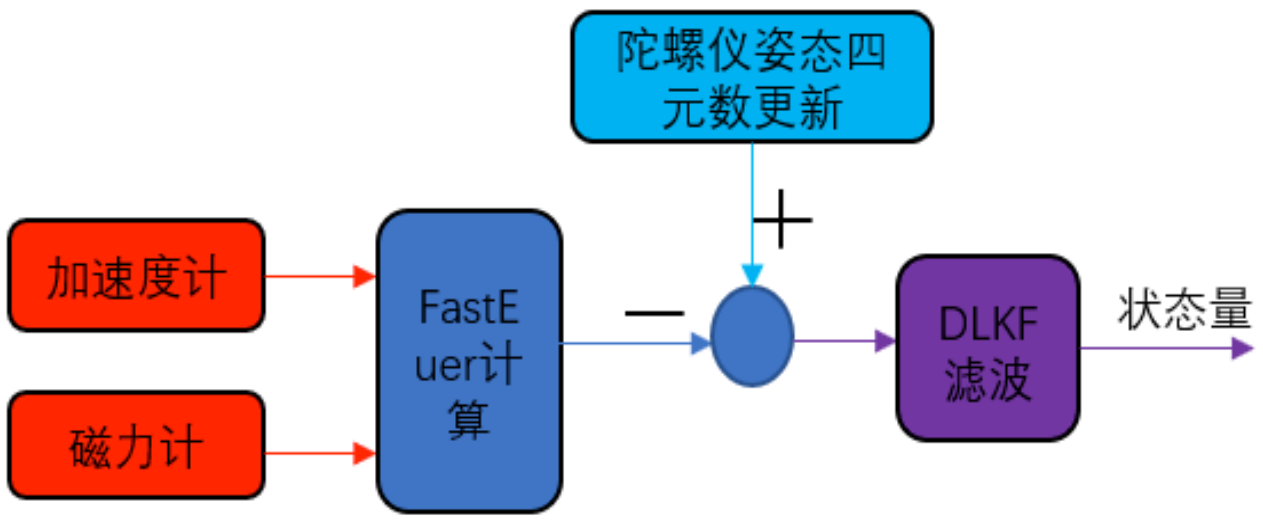

图 1 FastEuler-DLKF 航姿滤波算法框图

**算法 2：航姿组合滤波算法**

输入：gyro accel mag 以及滤波初值

输出：姿态角以及陀螺仪偏差

**While t > 0 do**

**Step 1: 陀螺仪姿态四元数更新**

**gyro = gyro – 陀螺仪偏差**

**quat = gyroUpdate(gyro,dt)**

**Step 2: 四元数转为姿态角**

$(\phi_{gyro},\theta_{gyro},\psi_{gyro})=\mathrm{quatToatt(quat)}$

**Step 3: FastEuler 计算**

$(\phi_a,\theta_a,\psi_m)=\mathrm{FastEuler(accel,mag)}$

**Step 4: DLKF 时间更新**

$(X,P)=\mathrm{TimeUpdate}(X_0,P_0,\Phi,\mathrm{dt})$

**Step 5: DLKF 第一层滤波修正**

$Z_1=(\phi_a-\phi_{gyro},\theta_a-\theta_{gyro})$

$(X_1,P_1)=\mathrm{accelCorrect}(X,P,R_{t_k}^a,H_1,Z_1)$

**Step 6: DLKF 第二层滤波修正**

$Z_2=\psi_{\mathrm{m}}-\psi_{\mathrm{gyro}}$

$(X_2,P_2)=\mathrm{magCorrect}(X_1,P_1,R_{\mathrm{m}},H_2,Z_2)$

**Step 7: 四元数误差修正**

$\mathrm{quat}=\mathrm{quatCorrect}(\mathrm{quat},X_2)$

$X_2 = 0$

**End while**

## 3 软件与硬件系统设计

无人机飞控系统根据不同调用频率的任务列表结构，将系统控制模式更新、控制律运行、导航数据更新、遥控器输入处理、与地面站数据交互和 LED 状态灯更新等任务以设定的频率执行，保证系统有序运行，如图 2 为飞行系统软件框图。

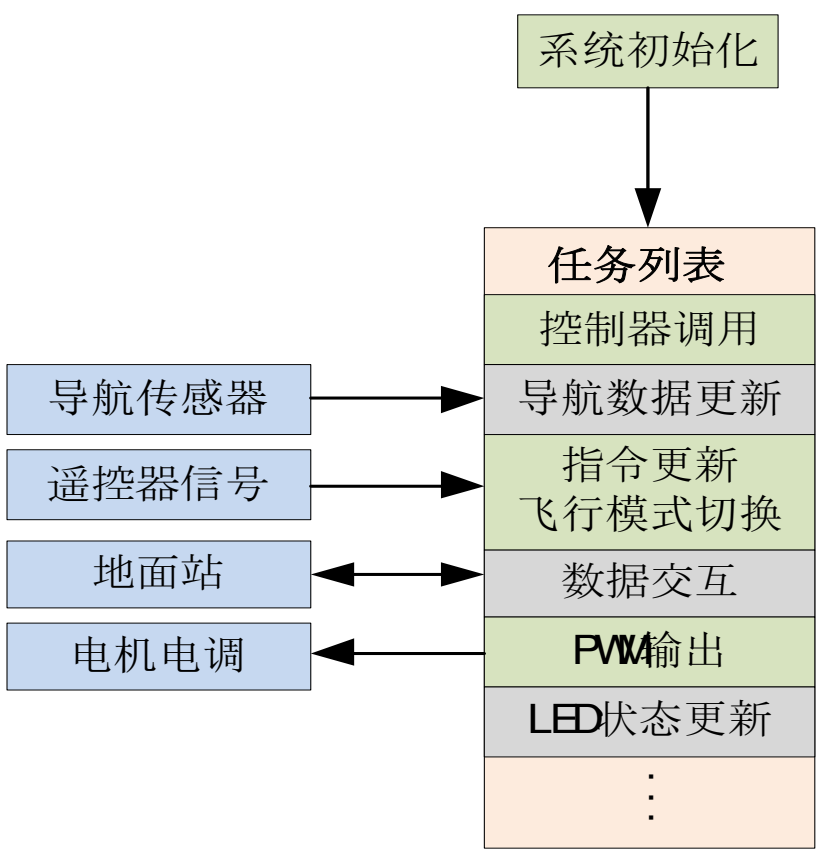

图 2 无人机飞行系统软件框图

本文算法在实验室自研飞控硬件上进行验证。

该型硬件主控芯片为 STM32F407， IMU 为带有三轴陀螺仪和三轴加速度计 MPU6050，磁力计选用内置 AK8975 和外置 HMC5883。根据自研飞控硬件及算法，在无人机平台上进行实际试飞，遥控器控制无人机并且通过数传将姿态数据实时发送给地面站，如图 3 所示为无人机硬件系统图。

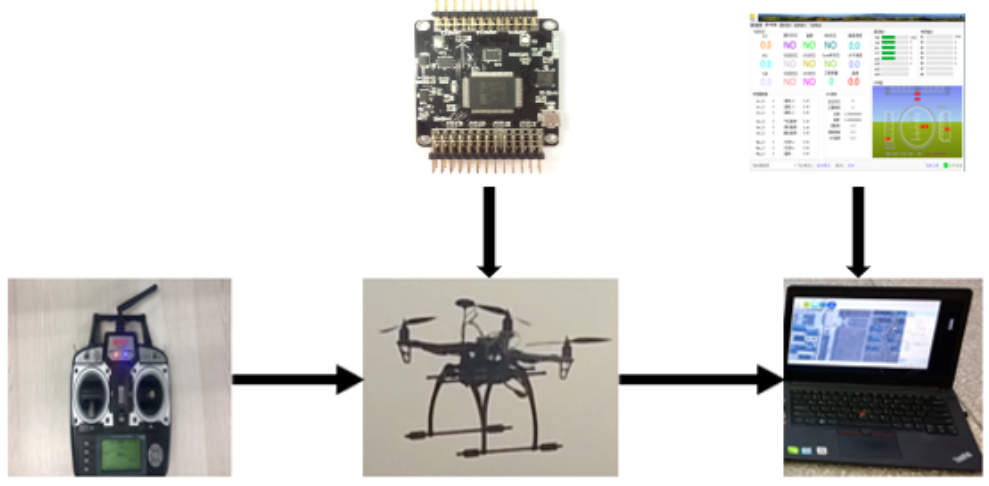

图 3 无人机硬件系统图

## 4 仿真与飞行测试

### 4.1 数据采集与飞行测试平台

本文算法数据仿真验证与飞行验证分别在两个不同的平台上进行。算法使用的仿真飞行数据是用图 4 的 Pixhawk 实验平台进行采集，将本文提出的航姿滤波算法在 matlab 软件上进行仿真，并且与互补滤波算法进行对比，验证算法的有效性与可行性。并且将算法嵌入到图 5 的试验平台，通过地面站实时观测姿态角曲线变化。

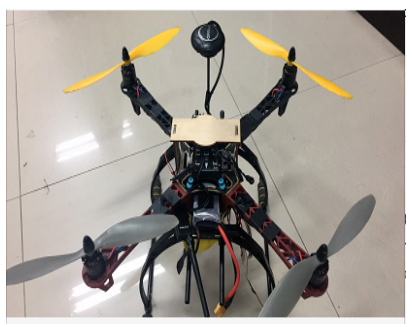

图 4 Pixhawk 实验平台

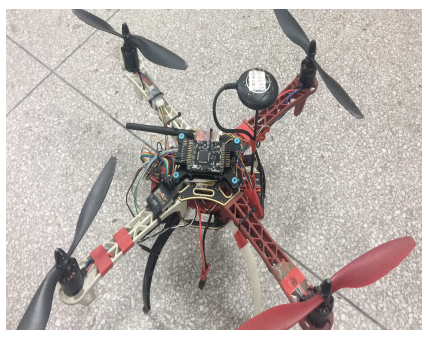

图 5 飞行测试平台

### 4.2 仿真实验

利用图 4 实验平台，无人机在定点模式下进行飞行数据的采集，并进行一些机动操作，图 6 是无人机飞行轨迹曲线图，导航坐标系是北-东-地，机体坐标系是前-右-下。

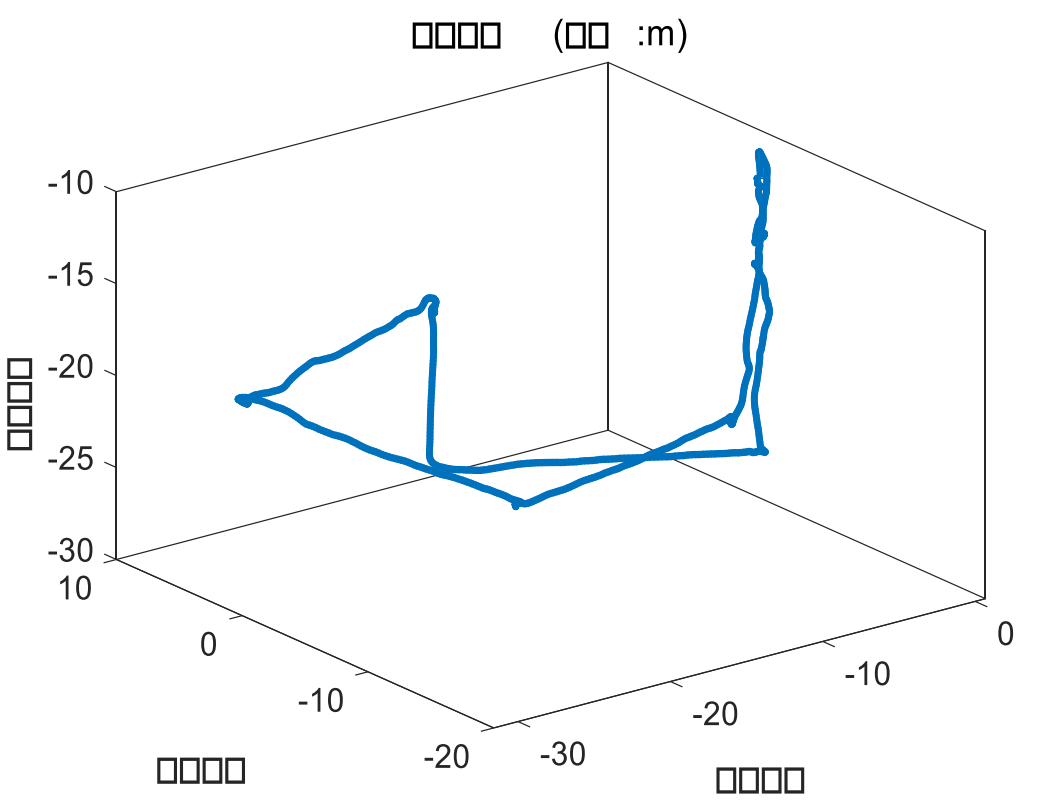

图 6 无人机飞行轨迹

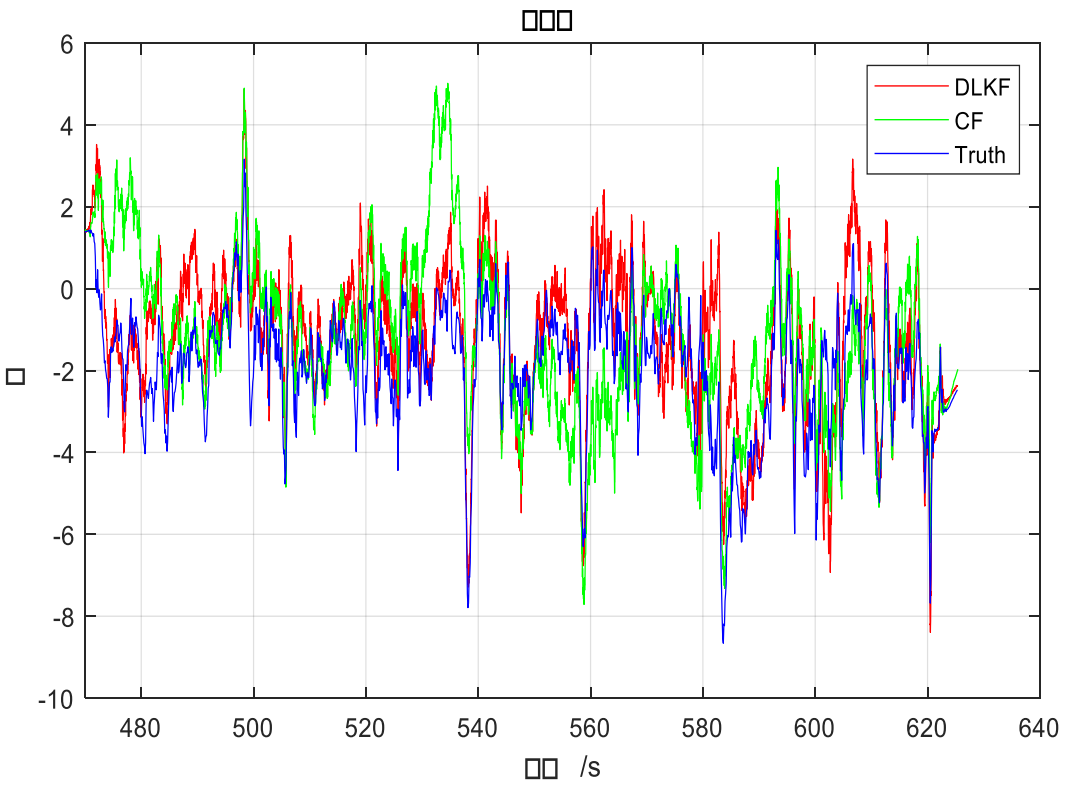

图 7 滚转角变化图

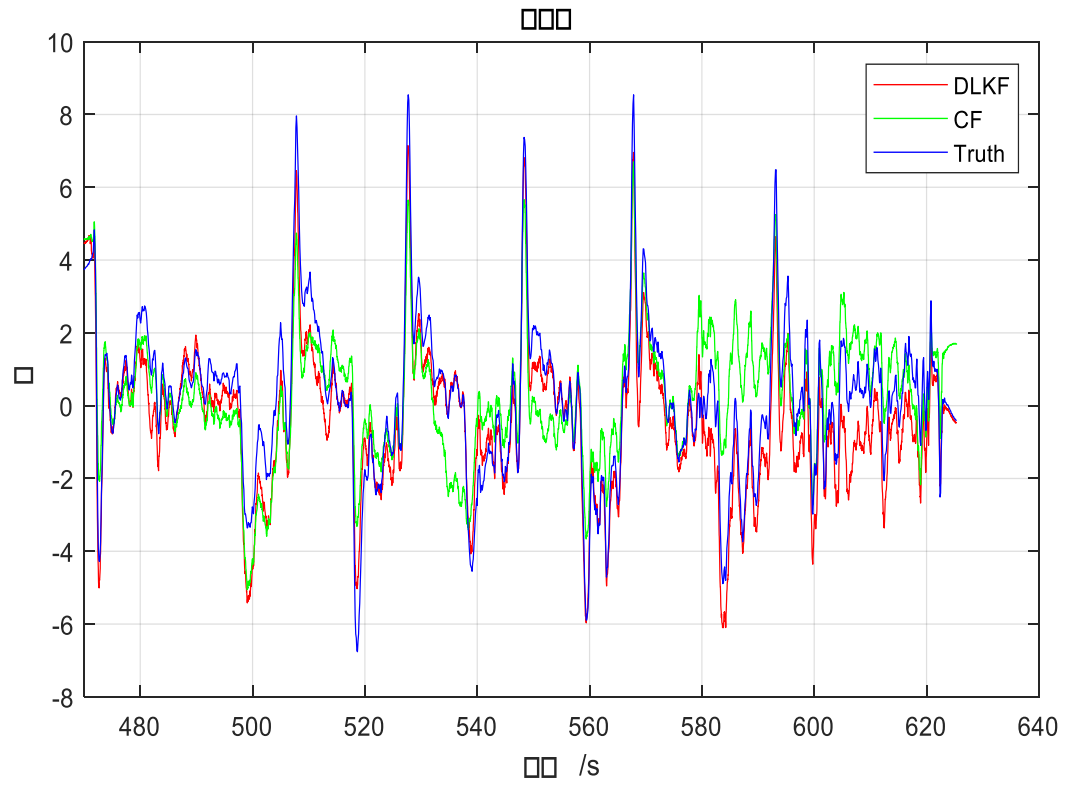

图 8 俯仰角变化图

图 7 和图 8 可以看出，如果以 Pixhawk 解算的姿态作为基准真值，DLKF 滤波解算的滚转角和俯仰角比互补滤波的解算精度更加准确，而且自适应因子的引入使得滤波具有鲁棒性，避免了互补滤波中出现的一些“解算突变”的情况。为了更加直观的进行对比，分别使用滚转角和俯仰角的均方根误差公式。

$$RMSE_{roll} = \sqrt{\frac{1}{N}\sum_{n=1}^{N}\left(\hat{\phi}_k^n - \phi_k^n\right)^2} \tag{24}$$

$$RMSE_{pitch} = \sqrt{\frac{1}{N}\sum_{n=1}^{N}\left(\hat{\theta}_k^n - \theta_k^n\right)^2} \tag{25}$$

**表 1 DLKF 和 CF 滚转角和俯仰角均方根误差**

| 滤波算法 | CF 滤波 | DLKF 滤波 |
|---|---|---|
| **Roll/度** | **1.7967** | **1.3156** |
| **Pitch/度** | **1.4317** | **1.0091** |

由表 1 可以看出，DLKF 滤波相比于 CF 滤波的解算精度，在滚转角中提高了大约 36.6%，在俯仰角中提高了大约 41.9%。

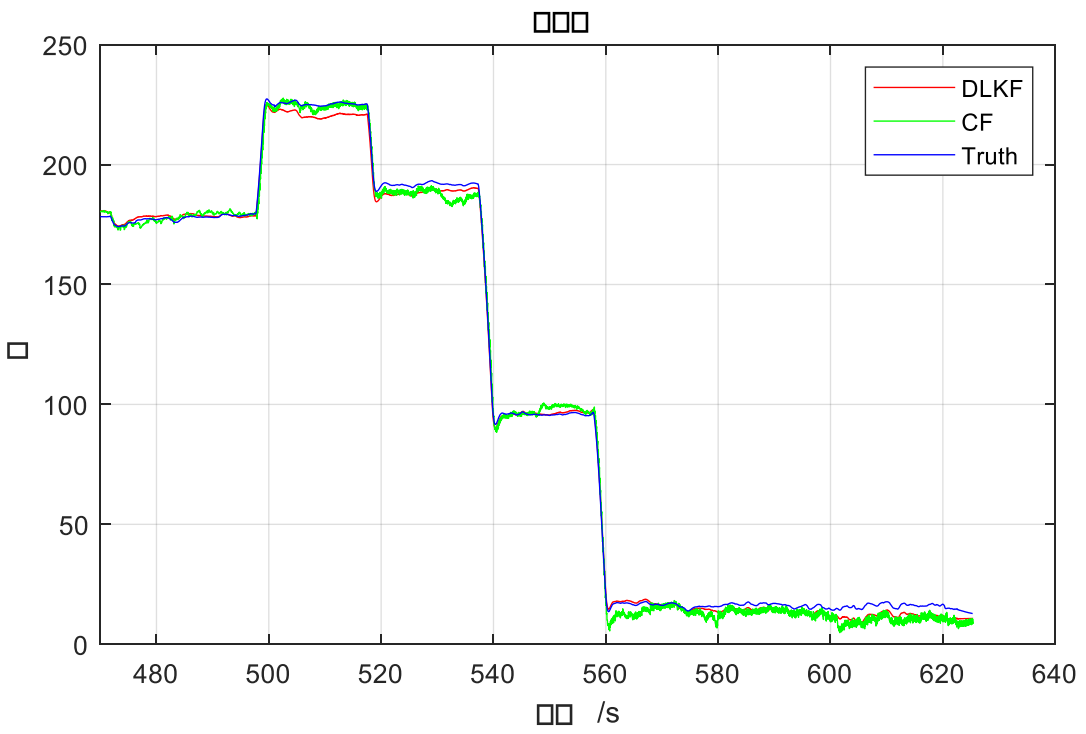

图 9 偏航角变化图

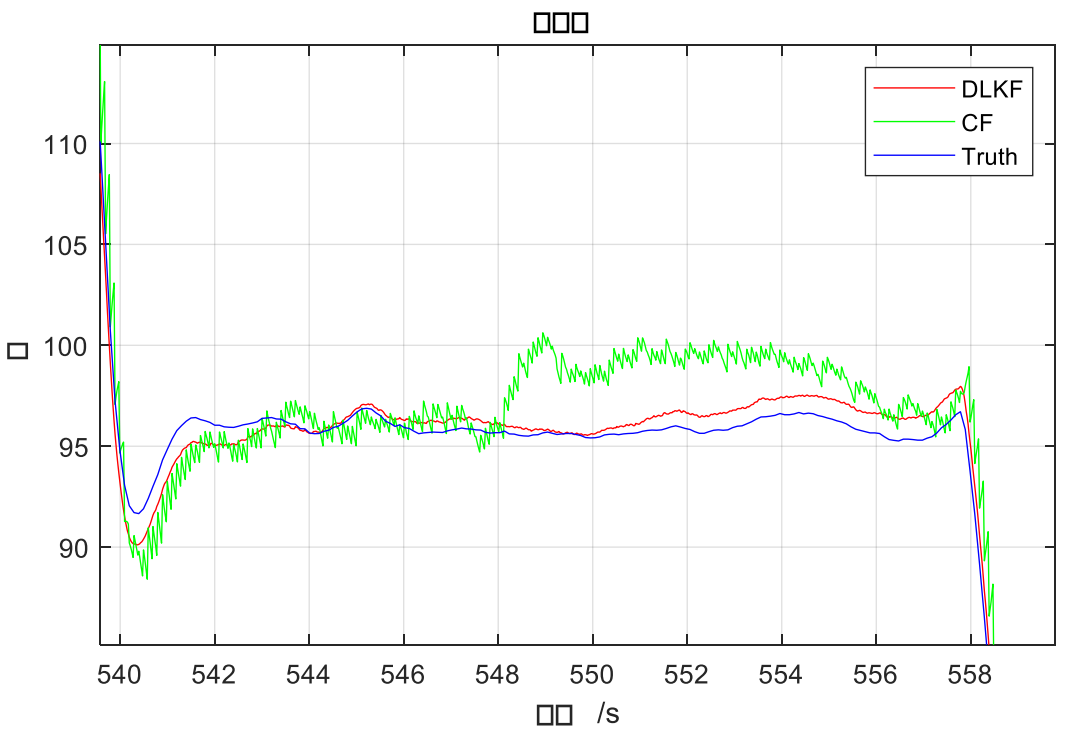

图 10 540s-558s 偏航角变化图

图 9 和图 10 可以看出，DLKF 滤波解算的偏航角与参考真值近似拟合。偏航角精度高的原因：一是 DLKF 滤波算法用磁力计进行修正，二是使用外置磁力计，减轻了电机转动产生的磁场对磁力计数据的影响。从图 9 中 CF 滤波的偏航角解算趋势虽然正常，但是放大 540s—558s 的变化趋势发现，CF 滤波解算曲线有明显地抖动现象，且曲线不平滑。为了更加直观的进行对比，使用偏航角的均方根误差公式。由表 2 可以看出，DLKF 滤波相比于 CF 滤波的解算精度，在偏航角中提高了大约 42.6%，

$$RMSE_{yaw} = \sqrt{\frac{1}{N}\sum_{n=1}^{N}\left(\hat{\psi}_k^n - \psi_k^n\right)^2} \tag{26}$$

**表 2 DLKF 和 CF 偏航角均方根误差**

| 滤波算法 | CF 滤波 | DLKF 滤波 |
|---|---|---|
| **Yaw/度** | **4.0636** | **2.850** |

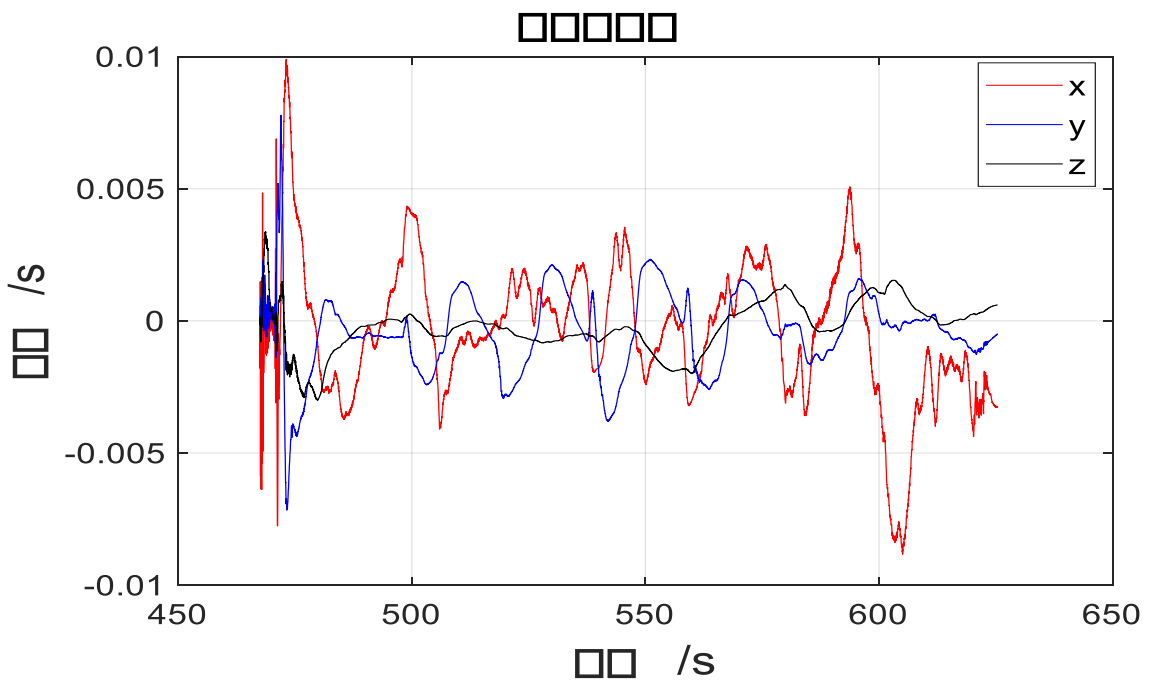

图 11 陀螺仪偏差变化图

图 11 估计的角速度偏差用于消除陀螺仪角速度时变随机误差，使角速度测量值精度更加准确。

## 4.3 飞行测试

利用图 5 自研飞控平台试验平台，将 4.2 小节验证的航姿算法通过 Keil Uvision5 开发环境编写嵌入式 C 程序，下载到自研飞控硬件，在地面站上实时显示飞行姿态曲线。

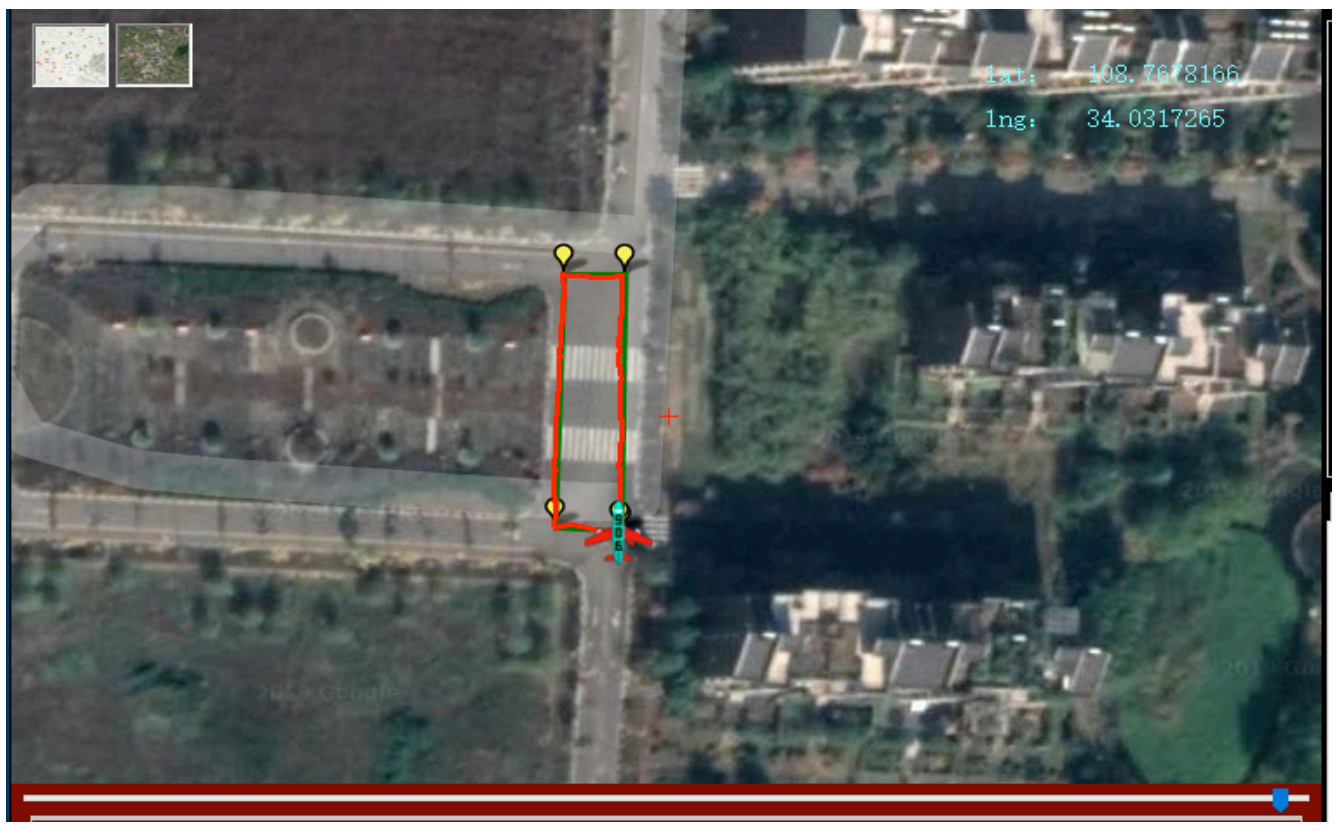

图 12 无人机飞行轨迹地图

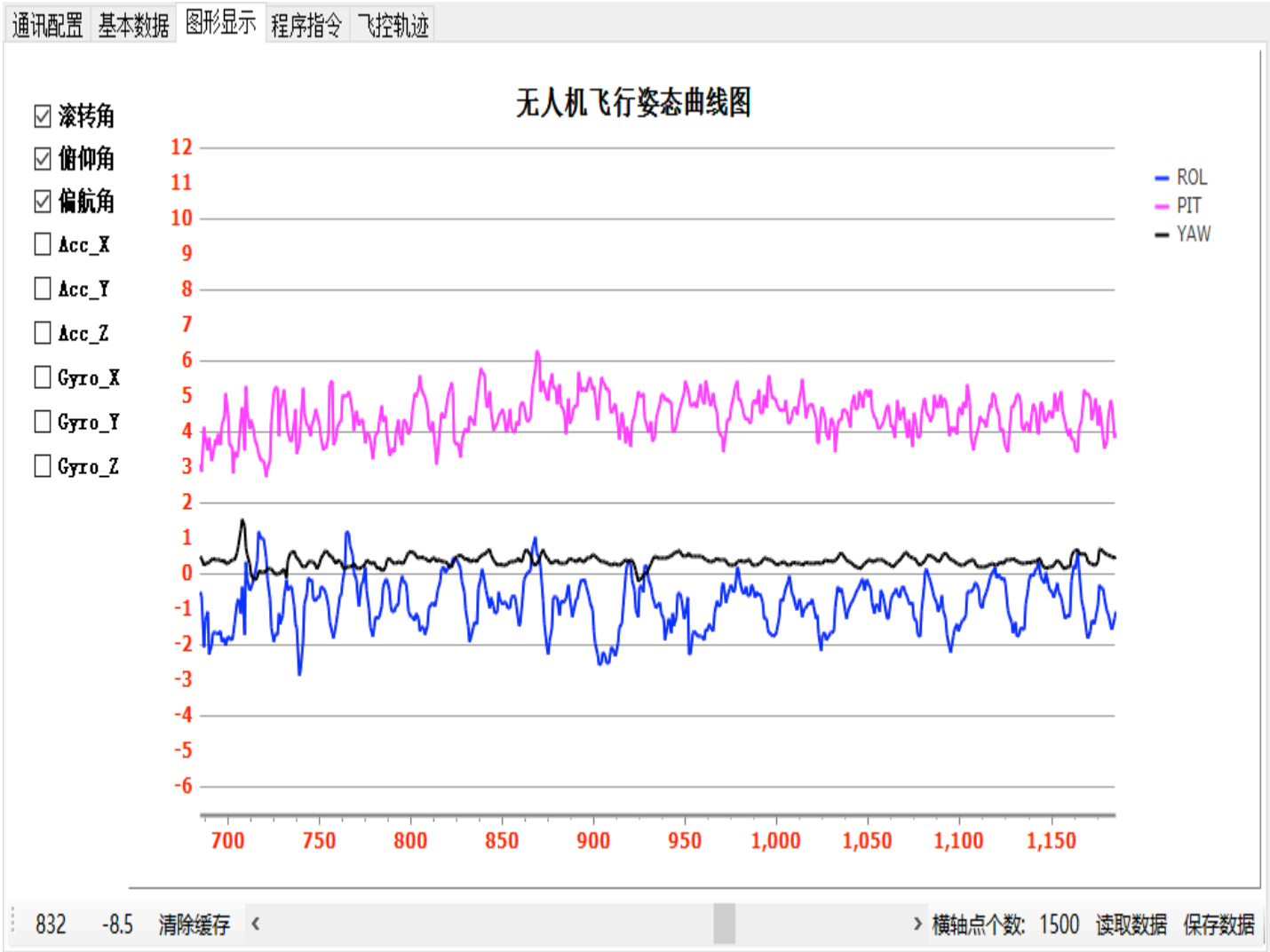

图 13 地面站飞行姿态曲线图

图 13 是无人机在预定的轨迹飞行时，通过地面站实时显示的姿态变化曲线。其中，飞行速度大约 0.5m/s，进行侧飞时，滚转角变化大约有 2 度，前飞时俯仰角变化大约 6 度。无人机在飞行时，机头近似指向正北方向且保持指向不变，以测试偏航角的解算稳定性。可以看出偏航角变化大约 1 度，解算精度可以满足小型无人机飞行要求。

## 5 结　论

本文针对小型无人机无人机航姿问题，提出了一种 FastEuler-DLKF 多传感器航姿滤波算法。分析了陀螺仪误差模型和姿态误差模型，并且简化了航姿组合模型，使得能够满足无人机姿态快速变化的应用场景。接着给出陀螺仪四元数姿态更新公式，

提出一种快速欧拉角计算方法并检测异常数值，并推导出航姿组合模型。详细介绍了 FastEuler-DLKF 航姿组合算法的整体运行流程。介绍了自研飞控硬件以及软件系统，利用 Pixhawk 实验平台和自研飞控试验平台，分别验证了算法的有效性及实际试飞的可行性。下一步工作将解决磁力计传感器出现故障时，如何进行故障检测以及与其他磁航向传感器进行融合估计，比如用 GPS 或者视觉解算航向角，使得航姿解算更加可靠和稳定。

## 参考文献（References）

[1] 陈宇捷. 基于MEMS的微小型嵌入式航姿参考系统研究[D]. 上海：上海交通大学，2009:5.
(Yujie Chen. Research on Micro-miniature Embedded Attitude Reference System Based on MEMS[D]. Shanghai: Shanghai Jiaotong University, 2009:5.)

[2] 刘琼. 微型航姿系统的设计与姿态解算算法研究[D]. 重庆：重庆邮电大学，2016:3-5.
(Qiong Liu. Research on Design and Attitude Algorithm of Micro Attitude System[D]. Chongqing: Chongqing University of Posts and Telecommunications, 2016:3-5)

[3] 袁政. 无人机航姿参考系统开发及信息融合算法研究[D]. 长沙：中南大学，2012:4-5.
(Zheng Yuan. Research on Development and Information Fusion Algorithm of UAV Attitude Reference System[D]. Changsha: Central South University, 2012:4-5.)

[4] Euston M, Coote P, Mahony R, et al. A complementary filter for attitude estimation of a fixed-wing UAV[C]//2008 IEEE/RSJ International Conference on Intelligent Robots and Systems. IEEE, 2008: 340-345.

[5] Wu, Jin, et al. "Fast complementary filter for attitude estimation using low-cost MARG sensors." IEEE Sensors Journal 16.18 (2016): 6997-7007.

[6] Madgwick, Sebastian OH, Andrew JL Harrison, and Ravi Vaidyanathan. "Estimation of IMU and MARG orientation using a gradient descent algorithm." 2011 IEEE international conference on rehabilitation robotics. IEEE, 2011.

[7] 吕印新. 基于 MEMS/GPS 的微型无人机组合航姿系统研究. MS thesis. 南京航空航天大学，2013.
(Yinxin Lu. Research on MEMS/GPS-based miniature UAV combined attitude and attitude system. MS thesis. Nanjing University of Aeronautics and Astronautics, 2013.)

[8] Welch, Greg, and Gary Bishop. "An introduction to the Kalman filter." (1995): 41-95.

[9] 付梦印，邓志红，张继伟. Kalman 滤波理论及其在导航系统中的应用[M]. 北京：科学出版社，2003:1-7.
(Mengyin Fu, Zhihong Deng, Jiwei Zhang. Kalman Filter Theory and Its Application in Navigation System[M]. Beijing: Science Press, 2003:1-7.)

[10] 赵琳，et al. "组合导航系统非线性滤波算法综述." 中国惯性技术学报 17.1 (2009): 46-52.
(Lin Zhao, et al. "Overview of Nonlinear Filtering Algorithms for Integrated Navigation Systems." Journal of Chinese Inertial Technology 17.1 (2009): 46-52.)

[11] 宋宇，翁新武，and 郭昕刚. "基于四元数 EKF 算法的小型无人机姿态估计." 吉林大学学报（理学版） 53.03 (2015): 511-518.
(Yu Song, Xinwu Weng, and Weigang Guo. "Azimuth Estimation of Small UAV Based on Quaternion EKF Algorithm." Journal of Jilin University (Science Edition) 53.03 (2015): 511-518.)

[12] Sabatelli S, Galgani M, Fanucci L, et al. A double-stage Kalman filter for orientation tracking with an integrated processor in 9-D IMU[J]. IEEE Transactions on Instrumentation and Measurement, 2012, 62(3): 590-598.

[13] 张树侠，何昆鹏. 陀螺仪性能参数表征与评定[J]. 导航与控制，2010，09(2).
Shuxia Zhang, Kunpeng He. Characterization and Evaluation of Gyro Performance Parameters[J]. Navigation and Control, 2010, 09(2).)

[14] 秦永元. 惯性导航[M]. 北京：科学出版社，2013:8,224-226,288-292,312,331,360.
(Yongyuan Qin. Inertial Navigation [M]. Beijing: Science Press, 2013:8,224-226,288-292,312,331,360.)

[15] Ljung, Lennart. "Asymptotic behavior of the extended Kalman filter as a parameter estimator for linear systems." IEEE Transactions on Automatic Control 24.1 (1979): 36-50.